\documentclass{article} % For LaTeX2e
\pdfoutput=1
\usepackage{nips15submit_e,times}
\usepackage{hyperref}
\hypersetup{pdfauthor={Jan Hendrik Metzen},pdftitle={Active Contextual Entropy Search}}
\usepackage{url}

\usepackage{graphicx}

\usepackage{amsmath,amssymb}
\def\D{\mathrm{d}}

\usepackage{umoline}

\title{Active Contextual Entropy Search}

\author{
Jan Hendrik Metzen \\
Universit{\"a}t Bremen, 28359 Bremen, Germany \\
DFKI GmbH, Robotics Innovation Center, 28359 Bremen, Germany \\
\texttt{jhm@informatik.uni-bremen.de} \\
}

\nipsfinalcopy % Uncomment for camera-ready version

\begin{document}

\maketitle

\begin{abstract}
Contextual policy search allows adapting robotic movement primitives to
different situations. For instance, a locomotion primitive might be adapted to
different terrain inclinations or desired walking speeds. Such an adaptation
is often achievable by modifying a small number of hyperparameters.
However, learning, when performed on real robotic systems, is typically
restricted to a small number of trials. Bayesian optimization
has recently been proposed as a sample-efficient means for contextual policy search
that is well suited under these conditions.
In this work, we extend entropy search, a variant of Bayesian optimization,
such that it can be used for \emph{active} contextual policy search where the agent
selects those tasks during training in which it expects to learn the most.
Empirical results in simulation suggest that this allows learning successful behavior with less trials.
\end{abstract}

\section{INTRODUCTION}

Contextual policy search (CPS) is a popular means for multi-task reinforcement
learning in robotic control \cite{deisenroth_survey_2013}. CPS learns a
hierarchical policy, in which the lower-level policy is often a
domain-specific behavior representation such as dynamical movement primitives
(DMPs) \cite{ijspeert_dynamical_2013}. Learning takes place on the upper-level
policy, which is a conditional probability density $\pi(\theta \vert
s)$ that defines a distribution over the parameter vectors $\theta$ of the
lower-level policy for a given context $s$. The context $s$ encodes
properties of environment or task such as a desired walking speed for a
locomotion behavior or a desired target position for a ball-throw behavior. The
objective of CPS is to learn an upper-level policy which maximizes the expected
return of the lower-level policy for a given context distribution.

CPS is typically based on local search based approaches such as cost-regularized
kernel regression \cite{Kober2012} and contextual relative
entropy search (C-REPS) \cite{kupcsik_data-efficient_2013,peters_relative_2010}.
From the field of black-box optimization, it is well-known that
local search based approaches are well suited for problems with a moderate
dimensionality and no gradient-information. However, for the special case of
relatively low-dimensional search spaces combined with an expensive cost
function, which limits the number of evaluations of the cost functions, global
search approaches like Bayesian optimization \cite{brochu_tutorial_2010} are often superior,
for instance for selecting hyperparameters \cite{snoek_practical_2012}. Combining contextual
policy search with pre-trained movement primitives\footnote{DMPs can
be pre-trained for fixed contexts in simulation or via some kind of imitation learning.}
can also fall into this category as
evaluating the cost function requires an execution of the behavior on the robot
while only a small set of hyperparameters might have to be adapted.
Bayesian optimization has been used for non-contextual policy search on
locomotion tasks \cite{calandra_bayesian_2015,lizotte_automatic_2007} and
robot grasping \cite{kroemer_combining_2010} and for (passive) contextual policy search
on a simulated robotic ball-throwing task \cite{metzen_bayesian_2015}.

In this work, we focus on problems where the agent can select the task
(context) in which it will perform the next trial during
learning. This facilitates active learning, which is considered to
be a prerequisite for lifelong learning \cite{ruvolo_active_2013}. A core
challenge in active multi-task robot control learning is the
incommensurability of performance in different tasks, i.e., how a learning
system can account for the relative (unknown) \emph{difficulty} of a task: for
instance, if a relatively small reward is obtained when executing a specific
low-level policy in a task, is it because the low-level policy is not well
adapted to the task or because the task is inherently more difficult than
other tasks?  Fabisch et al. \cite{fabisch_accounting_2015} presented an approach for
estimating the task-difficulty explicitly, which allows defining heuristic
intrinsic reward functions based on which a discounted multi-arm bandit selects the next task actively \cite{fabisch_active_2014}.

In this work, we follow a different approach: rather than explicitly addressing
the incommensurability of rewards, we propose ACES, an information theoretic
approach for active task selection which selects tasks not based on rewards
directly but rather based on the expected reduction in uncertainty about the
optimal parameters for the contexts. ACES allows selecting task and parameters
jointly without requiring a heuristic definition of a task selection criterion.
ACES is motivated by entropy search \cite{hennig_entropy_2012}, which has been
extended in a similar fashion to non-contextual settings \cite{alonso_an_2015},
and multi-task Bayesian optimization \cite{swersky_multi-task_2013}, which focuses on problems with  discrete context spaces.

\section{BACKGROUND}

\textbf{Contextual Policy Search} (CPS)
denotes a model-free approach to reinforcement learning,
in which the (low-level) policy $\pi_\theta$ is parametrized by a vector
$\theta$. The choice of $\theta$ is governed by an upper-level policy
$\pi_\omega$. For generalizing learned policies to multiple tasks,
the task is characterized
by a context vector $s$ and the upper-level policy
$\pi_\omega(\theta\vert s)$ is conditioned on the respective
context. The objective of CPS is to learn
$\pi_\omega$ such that the expected return $J_\omega$ over all contexts is
maximized, with $J_\omega = \int_s p(s) \int_\theta
\pi_\omega(\theta\vert s) R(\theta, s) \D{\theta
} \D{s}$. Here, $p(s)$ is the distribution over contexts and
$R(\theta, s)$ is the expected return when executing the low
level policy with parameter $\theta$ in context $s$. We refer to Deisenroth et al.
\cite{deisenroth_survey_2013} for a recent overview of (contextual) policy
search approaches in robotics.

\textbf{Bayesian optimization for contextual policy search} (BO-CPS)
is based on applying ideas from Bayesian optimization
to contextual policy search \cite{metzen_bayesian_2015}. BO-CPS learns internally a
model of the expected return $R(\theta, \mathbf{s})$ of a parameter vector $\theta$ in
a context $s$.  The model is based on Gaussian process (GP)
regression \cite{rasmussen_gaussian_2006}. It learns from sample returns $R_i$ obtained in rollouts at query points
consisting of a context $s_i$ determined by the environment and a parameter
vector $\theta_i$ selected by BO-CPS. By learning a joint GP model over the
context-parameter space, experience collected in one context is naturally
generalized to similar contexts.

The GP model provides both an estimate of the expected return $\mu_{GP}[R(s,
\theta)]$ and its standard deviation $\sigma_{GP}[R(s, \theta)]$.
Based on this information, the parameter vector for the given context is
selected by maximizing an \emph{acquisition function}, which allows controlling the trade-off between exploitation (selecting parameters with maximal
estimated return) and exploration (selecting parameters with high uncertainty).
Common acquisition functions used in Bayesian optimization such as the probability of
improvement (PI) and the expected improvement (EI) \cite{brochu_tutorial_2010}
are not easily generalized to BO-CPS \cite{metzen_bayesian_2015}.
In contrast, the acquisition function GP-UCB \cite{srinivas_gaussian_2010}, which defines the acquisition value of a parameter
vector in a context as $\text{GP-UCB}(s, \theta) = \mu_{GP}[R(s, \theta)] +
\kappa \sigma_{GP}[R(s, \theta)]$, where $\kappa$ controls the
exploration-exploitation trade-off, can be applied to BO-CPS straightforwardly resulting in an approach similar to CGP-UCB \cite{krause_contextual_2011}.
BO-CPS selects parameters $\theta_i$ for a given fixed context $s_i$ by performing an optimization over the parameter space using the global maximizer DIRECT \cite{jones_lipschitzian_1993} to find the approximate global maximum, followed by L-BFGS \cite{byrd_limited-memory_1995} to refine it.

\textbf{Entropy search} (ES) is a recently proposed approach to probabilistic
global optimization that mainly differs from Bayesian optimization in the
choice of the acquisition function \cite{hennig_entropy_2012}. While typical acquisition
functions used for Bayesian optimization select query points where they expect
the optimum, ES selects query points where it expects to learn most about the
optimum. More specifically, ES explicitly represents $p_{opt}(\theta)$, the
probability that the global optimum (maximum or minimum, depending on the
problem) of the unknown function $f$ is at $\theta$. ES estimates $p_{opt}(\theta)$
at finitely many points $\{\theta^c\}_{i=1}^{N_\theta}$ on a non-uniform grid
that are selected heuristically. Moreover, it approximates $p_{opt}$ at $\theta^c$ based
on expectation propagation or Monte Carlo integration. To select a query point,
ES predicts the change of the GP when drawing a sample at the query point
$\theta^q$ and assuming $N_y$ different outcomes $\{y^{(i)}\}$ sampled from the
GP's predictive distribution at $\theta^q$. Thereupon, ES selects a query point
which minimizes the average loss $\mathcal{L}(p_{opt}[\theta^q]) = - \int
p_{opt}[\theta^q](\theta) \log \frac{p_{opt}[\theta^q](\theta)}{U_I(\theta)}
\text{d}\theta$, i.e., which maximizes the relative entropy between $p_{opt}$
and a uniform measure $U_I$, where $p_{opt}[\theta^q]$ denotes the probability
distribution of the global optimum \emph{after} an assumed query at $\theta^q$.

\section{ACTIVE CONTEXTUAL ENTROPY SEARCH}

In this section, we present active contextual entropy search (ACES), an
extension of ES to CPS which allows selecting both parameters $\theta_q$ and
context $s_q$ of the next trial. Let $p_{max}(\theta \vert s)$ denote the
conditional probability distribution of the maximum expected return given context $s$ and let
the loss $\mathcal{L}^s(s^q, \theta^q) = \mathcal{L}(p_{max}[s^q,\theta^q](\theta \vert s)) - \mathcal{L}(p_{max}(\theta \vert s))$ denote the expected change of relative
entropy in context $s$ after performing a trial in context $s^q$ with parameter
$\theta^q$. A straightforward extension of ES to active learning in BO-CPS would
be selecting $s^q, \theta^q = \arg\min_{(s^q, \theta^q)} \mathcal{L}^{s^q}(s^q,
\theta^q)$, i.e., select the context $s^q$ in which the maximum increase of
relative entropy is expected. This, however, would not account for information gained about the
optima in contexts $s \neq s^q$ by a query at $(s^q, \theta^q)$.

ACES instead averages over the expected change in relative entropy at different
points in the context space: $\text{ACES}(s^q, \theta^q) = \sum_{i=1}^{N_s}
\mathcal{L}^{s^c_i}(s^q, \theta^q)$, where $\{s^c\}_{i=1}^{N_s}$ is a set of
contexts which is drawn uniform randomly from the context space. Unfortunately,
each evaluation of $\mathcal{L}^{s^c_i}$ is computationally expensive and thus
$N_s$ would have to be chosen small. On the other hand, GPs have an intrinsic
length-scale for many choices of the kernel and thus, a query in context $s^q$
will only affect $\mathcal{L}^{s^c_i}$ when $s^c_i$ is ``similar'' to $s^q$. We
define similarity between contexts based on the Mahalanobis distance $d_M(s^c_i,
s^q) =\sqrt{(s^c_i - s^q)S^{-1}(s^c_i - s^q)}$ with $S$ being a diagonal matrix
with the (anisotropic) length scales of the GP on the diagonal. Based on this we
can approximate $\text{ACES}(s^q, \theta^q) \approx \sum_{s \in \text{NN}(s^q,
\{s^c\}, N_{nn})}  \mathcal{L}^s(s^q, \theta^q)$ with NN returning the $N_{nn}$
nearest neighbors of  $s^q$ in $\{s^c\}$ according to the Mahalanobis distance.
A larger value of $N_{nn}$ corresponds to a better approximation of
$\text{ACES}(s^q, \theta^q)$ at the cost of a linearly increased computational
cost.

Candidate points $\theta^c(s)$ are selected by performing Thompson sampling on
$500$ randomly chosen $\theta$ with $N_\theta=20$ . The number of
trial contexts $N_s$ is set to $100$ and we compare empirically $N_{nn}=1$ and
$N_{nn}=20$. The quantity $p_{max}$ is approximated using Monte-Carlo
integration based on drawing 1000 samples from the GP posterior. The number of
samples from the GP's predictive distribution at $\theta_q$ for approximating
the average loss for a query point is set to $N_y=10$. Since there is noise in
the Monte Carlo estimates of $\mathcal{L}^s(s^q, \theta^q)$, we use CMA-ES
\cite{hansen_completely_2001} as optimizer rather than DIRECT.

\section{EVALUATION}

%Methods to be compared:
%  * BO-CPS-UCB
%  * BO-CPS-ES  [fixed context]
%  * BO-CPS-ACES_local [only take effect onto uncertaints in same task into account]
%  * BO-CPS-ACES_approx
%  * BO-CPS-ACES_full  [very slow]

We present results in a simulated robotic control task, in which the robot arm
COMPI \cite{COMPI} is used to throw a ball at a target on the ground encoded in
a two-dimensional context vector. The target area is $[1, 2.5]m \times [-1, 1]m$
and the robot arm is mounted at the origin $(0, 0)$ of this coordinate system. Thus, contexts
can be chosen from a two-dimensional context space: $s \in [1, 2.5] \times [-1, 1]$.
The low-level
policy is a joint-space DMP with preselected  start and goal angle for each joint and
all DMP weights set to 0. This DMP results in throwing a ball such that it hits
the ground close to the center of the target area. Adaptation to different
target positions is achieved by modifying a two-dimensional vector $\theta$:
the first component of $\theta$ corresponds to the execution
time $\tau$ of the DMP, which determines how far the ball is thrown, and the second
component to the final angle $g_0$ of the first joint, which determines the rotation of the arm
around the z-axis.

The upper-level policy\footnote{The upper-level policy could in principle be defined directly on the surrogate model. This would, however, require a computationally expensive maximization over the parameter space for each evaluation of the policy.} is a deterministic policy which selects parameters $\theta$ based on an affine function of context $s$. This policy is trained on the training data $\{(s_i, \theta_i, R_i)\}_i$ using the C-REPS policy update.
The limits on the parameter space are $g_0 \in \left[-\frac{\pi}{2}, \frac{\pi}{2}\right]$ and $\tau \in \left[0.4, 2\right]$. All approaches use an anisotropic Mat{\'e}rn kernel for the GP surrogate model.
Since we focus on a ``pure exploration'' scenario \cite{bubeck_pure_2009}, GP-UCB's exploration parameter $\kappa$ is set to a constant value of $5.0$. The reward is defined as
$r = -||s - b_s||^2 - 0.01 \sum_t v_t^2$,
where $s$ denotes the goal position, $b_s$ denotes
the position hit by the ball, and $\sum_t v_t^2$ denotes a penalty term on the sum of squared joint velocities during DMP execution. The maximum achievable reward for different $b_s$ differs as values of $b_s$ further away from the origin (where the arm is mounted) require larger joint velocities $v_t$ and thus incur a larger penalty. Thus, rewards in different contexts are incommensurable.

\begin{figure}
\centering
\includegraphics[width=0.58\textwidth]{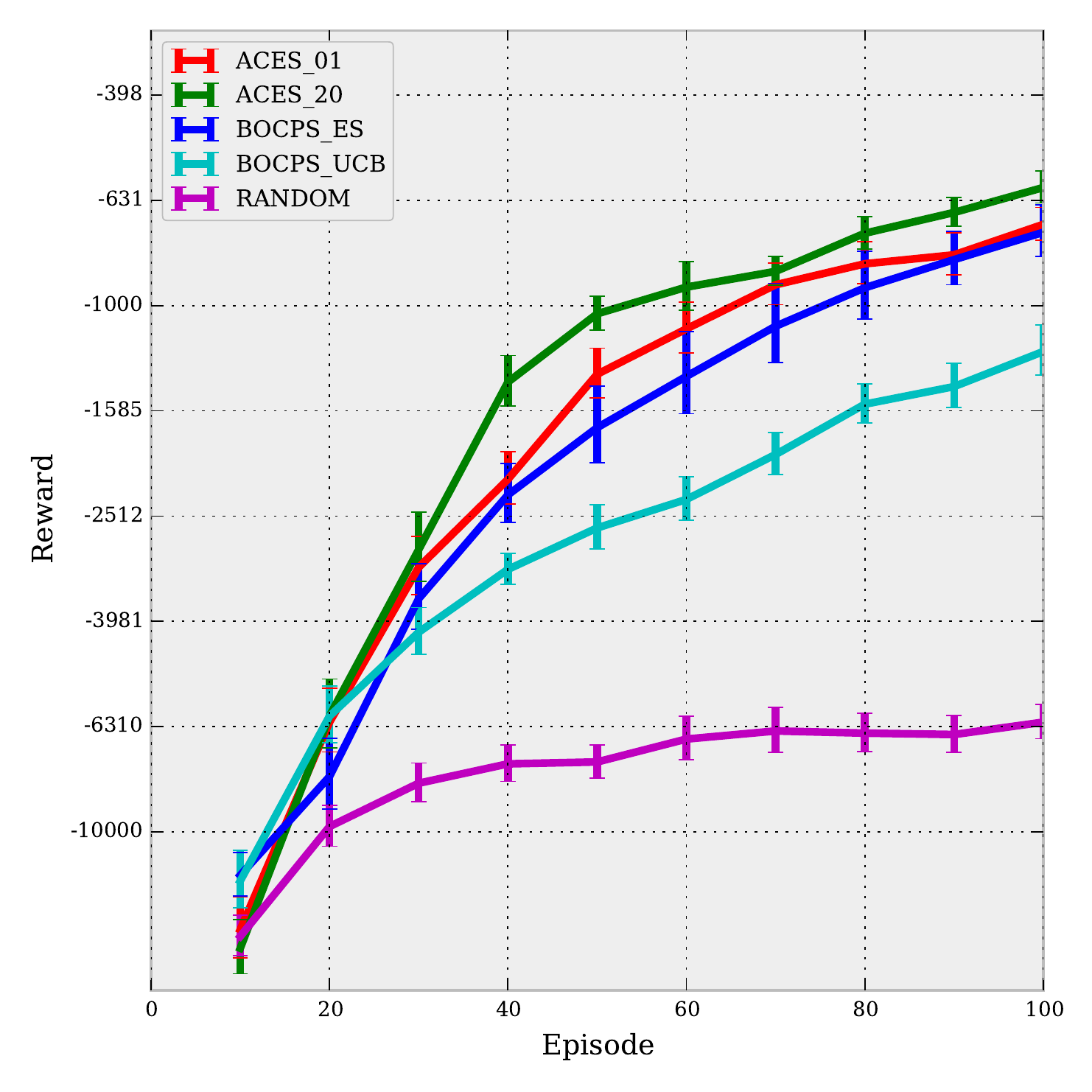}
\includegraphics[width=0.4\textwidth]{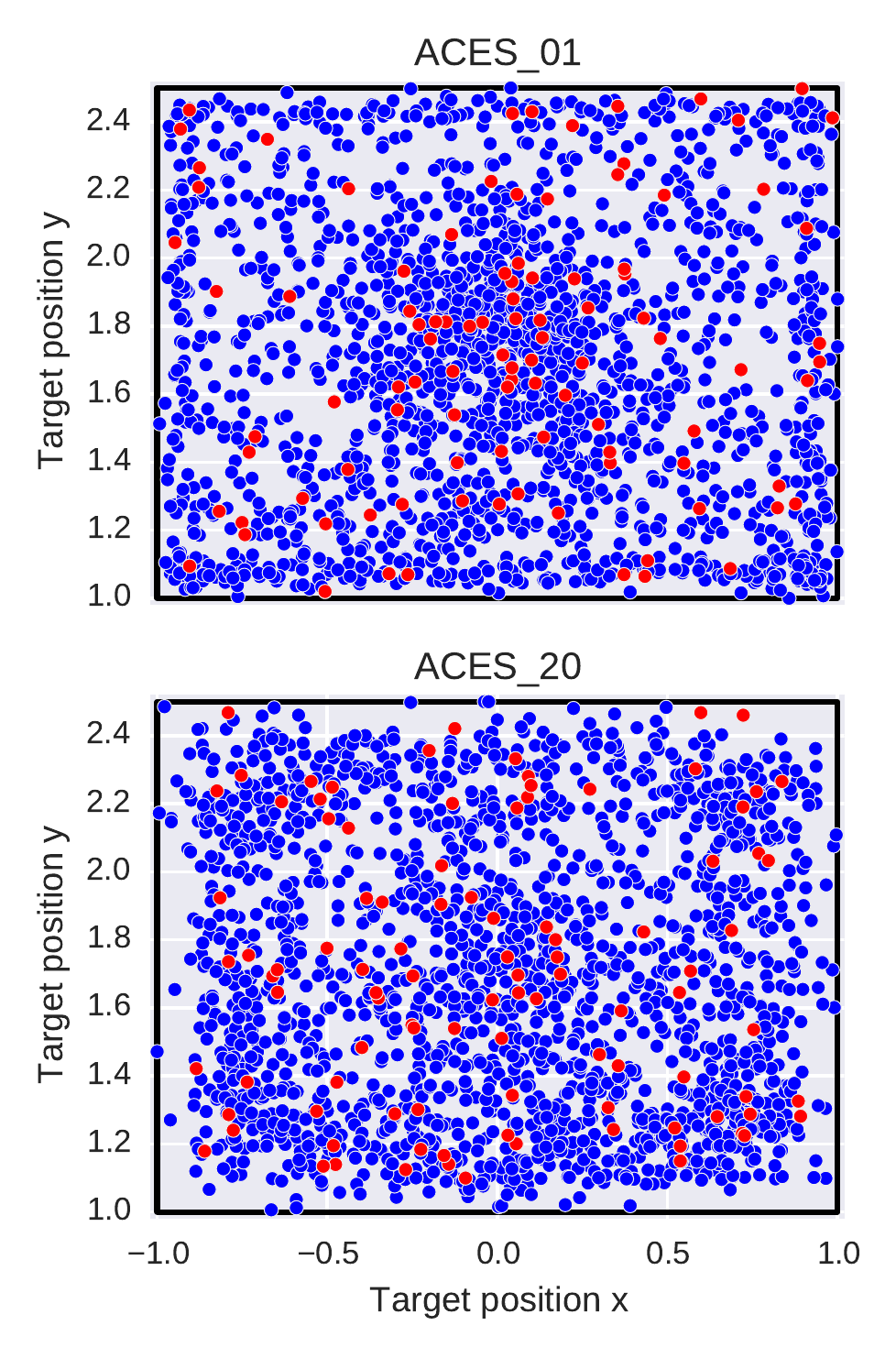}
\caption{(Left) Learning curves on the simulated robot arm COMPI: the offline performance is evaluated each 10 episodes on 16 test contexts distributed on a grid over the target area. Shown are mean and its standard error over 20 independent runs. (Right) Scatter plot showing the sampled contexts for all (blue) and a single representative run (red).}
\label{fig:experiment}
\end{figure}

Figure \ref{fig:experiment} summarizes the main results of the empirical
evaluation. The left graph shows the mean offline performance of the
upper-level policy at 16 test contexts on a grid over the context space.
Sampling contexts and parameters randomly during learning (``Random'') is shown
as a baseline and indicates that generalizing experience using a GP model alone
does not suffice for quick learning in this task. Rather, a non-random way of
exploration is required. BO-CPS with random context selection and UCB for
parameter selection improves considerably over random parameter selection. Using
ES for parameter selection further improves the learning speed. Closer
inspection (not shown) indicates that ES improves over UCB mainly because UCB
samples often at the boundaries of the parameter space since the uncertainty is
typically large there. ES samples more often in the inner regions of the
parameter space since those regions promise a larger information gain globally.

Active context selection using ACES further improves over BOCPS-ES, in
particular when the sum over the context space is approximated using
several samples ($N_{nn} = 20$ in the case of ACES\_20) rather than a single sample
($N_{nn} = 1$ for ACES\_01). The right graph shows the contexts selected by different variants of
ACES. It can be seen that ACES\_20 avoids selecting targets close to the
boundary of the context space as those typically reveal less global information
about the context-dependent optima as boundary points are far away from most
other regions of the context space. We attribute the improved learning
performance to this way of selecting targets during learning. In contrast,
ACES\_1 samples more often close to the boundaries as it only considers the local information gain
and thus has no reason to prefer inner over boundary contexts.

\section{DISCUSSION AND CONCLUSION}

We have presented an active learning approach for contextual policy search based
on entropy search. First experimental results indicate that the proposed active
learning approach provides considerable speed-ups of the learning of movement
primitives compared to a random task selection. Comparison with other active task selection approaches \cite{fabisch_active_2014} remains future work. Moreover, investigating and enhancing the scalability to higher dimensional problems, potentially by employing a combination with random embedding approaches such as REMBO
\cite{wang_bayesian_2013}, and combining active task selection with predictive entropy search \cite{hernandez-lobato_predictive_2014} or portfolio-based approaches \cite{shahriari_entropy_2014} would be interesting.

%Future work:
% * compare with methods like GP-REPS/PILCO?
% * make it on-policy

\paragraph{Acknowledgments}
This work was performed as part of the project \texttt{BesMan}\footnote{More information are available at \url{http://robotik.dfki-bremen.de/en/research/projects/besman.html}.} and supported through two grants of the German Federal Ministry of Economics and Technology (BMWi, FKZ 50 RA 1216 and FKZ 50 RA 1217).

\bibliographystyle{abbrv}
\bibliography{literature}

\end{document}